\documentclass[10pt,twocolumn,letterpaper]{article}

\usepackage[pagenumbers]{wacv} 

\usepackage{graphicx}
\usepackage{amsmath}
\usepackage{amssymb}
\usepackage{booktabs}

\usepackage{makecell}
\usepackage{pifont}
\usepackage{array}
\usepackage{textcomp, gensymb}
\usepackage{multirow}

\usepackage[accsupp]{axessibility}  

\usepackage[pagebackref,breaklinks,colorlinks]{hyperref}

\usepackage[dvipsnames]{xcolor}

\usepackage[capitalize]{cleveref}
\crefname{section}{Sec.}{Secs.}
\Crefname{section}{Section}{Sections}
\Crefname{table}{Table}{Tables}
\crefname{table}{Tab.}{Tabs.}

\def\wacvPaperID{14} 
\def\confName{WACV}
\def\confYear{2024}

\begin{document}

\title{Learning to Read Analog Gauges from Synthetic Data}

\author{Juan Leon-Alcazar\\
KAUST\\
Thuwal, Saudi Arabia\\
{\tt\small juancarlo.alcazar@kaust.edu.sa}
\and
Yazeed Alnumay\\
Aramco\\
Thuwal, Saudi Arabia\\
{\tt\small yazeed.alnumay@aramco.com}
\and
Cheng Zheng\\
KAUST\\
Thuwal, Saudi Arabia\\
{\tt\small cheng.zheng@kaust.edu.sa}
\and
Hassane Trigui\\
Aramco\\
Thuwal, Saudi Arabia\\
{\tt\small hassane.trigui@aramco.com}
\and
Sahejad Patel\\
Aramco\\
Thuwal, Saudi Arabia\\
{\tt\small sahejad.patel@aramco.com}
\and
Bernard Ghanem\\
KAUST\\
Thuwal, Saudi Arabia\\
{\tt\small bernard.ghanem@kaust.edu.sa}
}

\maketitle

\begin{abstract}
    Manually reading and logging gauge data is time-inefficient, and the effort increases according to the number of gauges available. We present a computer vision pipeline that automates the reading of analog gauges. We propose a two-stage CNN pipeline that identifies the key structural components of an analog gauge and outputs an angular reading. To facilitate the training of our approach, a synthetic dataset is generated thus obtaining a set of realistic analog gauges with their corresponding annotation. To validate our proposal, an additional real-world dataset was collected with 4.813 manually curated images. When compared against state-of-the-art methodologies, our method shows a significant improvement of 4.55$^{\circ}$  in the average error, which is a 52\% relative improvement. The resources for this project will be made available at: \url{https://github.com/fuankarion/automatic-gauge-reading}.
    \vspace{-0.5cm}
\end{abstract}


\section{Introduction}
\label{sec:introduction}

Analog gauges are widespread across modern industrial facilities. Typically, these gauges monitor critical operational parameters such as temperature, pressure, and level indicators for active industrial processes. Therefore, keeping an accurate record of gauge reading data is essential to track an asset's trends and its conditions over time, which would aid in failure investigation if one occurs.  

Manually reading and logging gauge data is time-inefficient, and increases according to the number of gauges available. To approach this problem, digital transmitters allow the remote monitoring of gauges. However, in many cases, the facilities are aged, or it is prohibitively expensive to upgrade numerous analog gauges to digital ones. Additionally, on-site readings of gauges serve as a direct verification means for validating the accuracy of digital transmitters, thus ensuring redundancy and process safety. 

\begin{figure}[t]
    \includegraphics[width=0.95\linewidth]{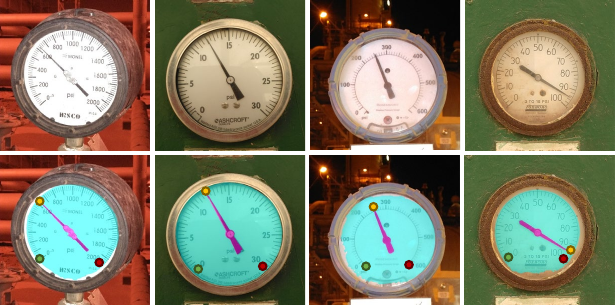}
    \caption{\textbf{Sample Results.} The top row shows samples of gauges captured on the field. The bottom row displays our model's predicted segmentation and regression of landmarks in these images. The segmentation result has three classes: background (not highlighted for easier visualization), gauge (cyan), and needle pointer (magenta). The gauge overlay also contains circles which visualize the estimated locations of the {\color[rgb]{0.44,0.68,0.28}• start marker}, {\color[rgb]{1,0.75,0}• needle pointer}, and {\color[rgb]{0.75,0,0}• end marker}.}
    \label{fig:field_seg}
    \vspace{-0.2cm}
\end{figure}
 
Since on-site readings can not be completely removed from the verification pipeline, some research efforts have explored the automatic reading of gauges from image data or video data captured directly in the facility \cite{alegria2000automatic, lauridsen2019reading}. These studies are largely limited by the availability of datasets with real-world data and curated ground-truth annotations. Unlike most modern machine learning datasets \cite{deng2009imagenet, lin2014microsoft, kay2017kinetics}, analog gauge reading data is still small and insufficient to train deep convolutional models. This data deficiency can be explained as it is expensive to collect and manually label a large number of real-world gauges. Moreover, privacy and security issues arise if these gauges are actively in use in an industrial facility. 

In this paper, we approach the automatic analog gauge reading task and formulate a  computer vision pipeline that enables the direct reading of analog gauges by processing a single picture of the corresponding device. Thereby providing automated readings which facilitates the digitization process, and simplifies the operator intervention. Our approach focuses on generating synthetic gauge images as a surrogate training set, and formulating a training time strategy that enables generalization from synthetic training data into real-world gauges. \Cref{fig:field_seg} contains some sample results from our method applied to real-world gauges.

We construct a synthetic gauge dataset by leveraging 3D modeling software. A single 3D model allows us to render images of various types of gauges from different view angles and simulates common noise sources. While synthetic gauge modeling mitigates the scarcity of training data, it poses a new challenge as it creates a domain gap between the rendered data (training) and the real-world examples (validation). With careful image augmentation, this domain gap can be effectively reduced, resulting in a model that can successfully operate in real-world scenarios. We validate the performance of our proposal on two real-world datasets containing over 7.500 labeled images.

Our paper brings 3 contributions. i) We develop a Blender model that simulates analog gauges and can render gauge images that contain the spatial location of the gauge's main components, and its reading. ii) We propose a two-stage (attention and classification) pipeline that identifies the key structural elements of the gauge and outputs a final reading. Paired with our training data this two-step approach produces state-of-the-art results. iii) We create a real-world dataset with 4,813 gauge readings which are manually curated and include multiple capture conditions and noise sources.

\section{Related Work}
\label{sec:related_work}

Automatic analog gauge reading is a machine vision task that involves object detection and landmark regression \cite{Howells_2021_CVPR}. Despite having a well-defined structure and goal, its solution is non-trivial. Therefore, multiple research efforts have addressed this problem. Overall, these approaches can be split into two main categories: classical computer vision and Deep Learning methods. 

Most of the traditional computer vision methods \cite{alegria2000automatic, jinuntuya2023coordinates, lauridsen2019reading, ma2018robust, tian2019pointer, wang2018automatic} rely on the shape of the gauge (typically circular), and apply the Hough Transform \cite{4767964} to estimate an initial gauge location. Upon this initial step, most of the approaches follow a secondary detection step which aims at establishing the location of key landmarks inside the gauge, commonly the pointer needle or the start and end markers \cite{Howells_2021_CVPR}. The work of Bao \textit{et al.} \cite{bao2019computer} uses adaptive thresholding to enable a second Hough transform step that approximates the location of the pointer needle. The approach of Yi \textit{et al.}\cite{yi2017clustering} proposed to use K-means clustering to identify the gauge elements, a similar method is outlined in \cite{lauridsen2019reading} where PCA and clustering are used to estimate the needle location on a binary image. The needle location step has also been approached by template matching and hand-crafted feature extraction (SIFT and RANSAC) \cite{8492946}. Finally, some works have analyzed this task in video streams \cite{9813954} and rely on the Optical Flow for temporal stability on the predictions.

Following the success of Deep Learning methods for natural image processing \cite{krizhevsky2017imagenet, he2016deep}, recent approaches have started training deep models that can directly regress the readings on a gauge. Some implementations rely on Deep Networks only for the detection procedure, commonly a CNN tuned for gauge detection \cite{9211895, 9960268}, and apply a Neural Network trained for Optical Character Recognition in order to parse the scale values \cite{9211895, li2020high}. 

Currently, most of these Deep Learning approaches suffer from a lack of training data and can easily overfit the visual appearance of the gauges in the training set\cite{lin2020pointer, liu2020detection}. This lack of generalization also holds for deep models that were trained on hybrid data, where the pointer needles of real gauges are artificially rotated \cite{cai2020pointer, 9960268}. Recently, \cite{Howells_2021_CVPR} devised a four-keypoint detection model to regress the minimum and maximum markers and the needle center and tip locations. Their model was trained on a large fully synthetic dataset and showed improved generalization to real images.

In contrast to the current approaches, we devise a simple, yet effective deep-learning pipeline with only two-trainable stages. We show that this strategy can be trained exclusively on synthetic data and achieve state-of-the-art results on real-world datasets.

\section{Automatic Gauge Reading}
\label{sec:method}

\begin{figure*}[t]
    \centering
    \includegraphics[width=\linewidth]{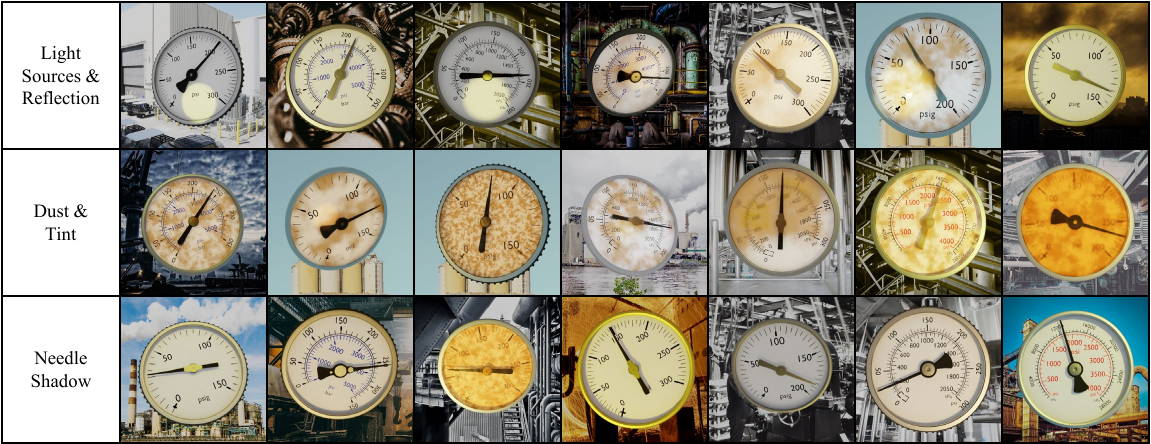}
    \caption{ \textbf{Synthetic Gauge Dataset.} We use the 3D creation suite Blender to generate diverse synthetic training samples. We design the rendered samples to include common artifacts and sources of noise, which makes the automatic gauge reading task more challenging. For example, we simulate light reflections on the gauge crystal and case (first row). We can include different patterns of dust/sand coverage over the crystal (second row). We also control for the presence of the needle's shadow, which can generate alignment errors in the landmark detection task (third row).}
    \label{fig:synth_samples}
    \vspace{-0.2cm}
\end{figure*}

In this section we outline the 3 stages of our approach: i) synthetic training data, ii) gauge component location, and iii) gauge landmark reading. The first component generates surrogate training data, by rendering synthetic gauges from a 3D model. The last two components are instantiated as convolutional neural networks \cite{lecun1989backpropagation, lecun1998gradient, krizhevsky2017imagenet}. These networks aim at estimating a fine-grained label of all the gauge components, and then approximate the location of 3 key landmarks in the gauge: start marker, end marker, and the pointer needle's tip.

\subsection{Synthetic Training Data}
\label{subsec:synthetic_dataset}
We decide against collecting training samples from gauges and manually labeling their readings. Such a process is expensive and slow and can be difficult to implement given the security constraints in some industrial facilities. Additionally, after gathering and labeling a large amount of data, we would still have to assure that it contains visually diverse gauges, and includes multiple capture conditions. Such diversity would mitigate the risk of overfitting to the global visual appearance of the collected gauges, instead of approximating the key landmarks on it. 

For the training phase, we rely on synthetic data that approximates the visual appearance of an analog gauge. We name this data split the \textit{synthetic dataset}. We approach this task with the open-source 3D creation suite Blender \cite{blender}. In Blender, we design a 3D model of a circular gauge which allows us to manipulate key elements of its appearance like needle type, case style, sticker min and max values, size, tint on the front crystal, and simulated dust particles amongst many others. \Cref{fig:synth_samples} contains some samples of the synthetic dataset.

The flexibility of Blender allows us to freely mix and manipulate all the model elements in order to create multiple unique synthetic gauges. Moreover, we can adjust the camera parameters and lighting patterns while rendering an arbitrary view of the synthetic gauge. Finally, we can also control the needle orientation, which allows us to simulate multiple ground-truth values. This flexibility enables us to train convolutional networks in a fully supervised manner, since the ground-truth location of all the gauge components and the landmark angular locations are generated along with the rendered training data. 

We build a Blender script to randomize all gauge attributes and render the training dataset. The resulting synthetic dataset contains 12,000 images. Along with the rendered image, the script generates the corresponding ground-truth segmentation mask with 3 classes (background, gauge, needle), the angular location of the needle's tip, and the maximum and minimum markers of the gauge sticker. 
 
\subsection{Gauge Detection and Reading}
Our approach for automatic gauge reading is a two-step classification pipeline. We first establish an approximate location of the most relevant gauge physical components, this estimation is then used as a spatial guidance to a second stage that effectively estimates the location of the key landmarks in the gauge, thus enabling a final angular reading.

\begin{figure*}[t]
    \centering
    \includegraphics[width=0.925\linewidth]{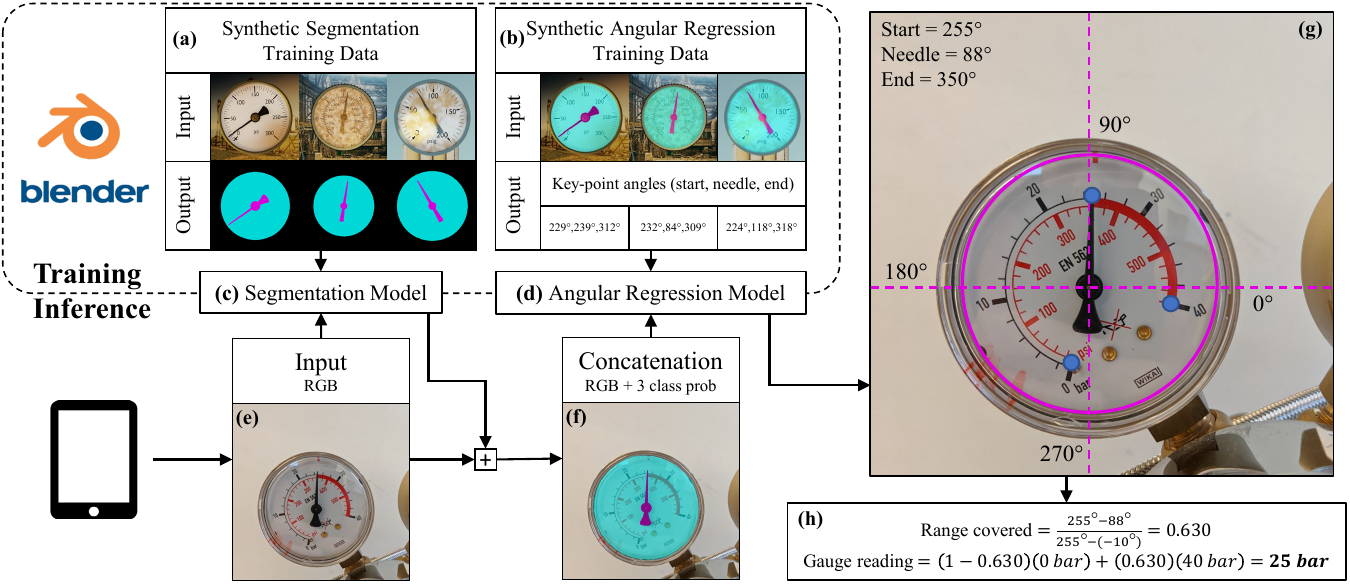}
    \caption{\textbf{Method Overview.} We generate  synthetic training data with diverse appearances and levels of noise from the Blender component (a). Additionally, we render the ground-truth semantic segmentation and output the location of the gauge key-points For all the synthetic data (b). At training time, we first optimize a segmentation network (c) that provides a localization prior for the gauge components. Then we use a concatenated tensor of the synthetic image and its estimated segmentation mask (f) to train a reading network that approximates the angular location of the pointer needle and the sticker start and end points (d). The angular locations of these points along with the known gauge unit and range are used to determine the gauge's reading (g, h).}
    \label{fig:model_pipeline}
    \vspace{-0.2cm}
\end{figure*}

In practice, the first step is approached by performing semantic segmentation on the gauge images, we estimate a pixel-wise mask of the gauge, thus partitioning it into 3 classes: Gauge Case, Needle, and Background. The second step is approached as a classification problem, where we establish the angular location of three key-points in the gauge: start marker, end marker, and pointer needle location. The combination of these 3 key-points provides an estimation of the position of the needle relative to the start and end of the gauge scale.

\vspace{-0.3cm}
\paragraph{Segmentation Network}
In the location step, we label every individual pixel in the input image. As a result, we obtain an approximate spatial location of the gauge in the picture (given by the gauge case class), and an initial location of the most relevant landmark in the process, namely the pixels assigned to the needle class. We approach the segmentation step using DeepLabv3 \cite{deeplabV3+}, with a ResNet-18 backbone \cite{ResNet}. At training time, we rely exclusively on the synthetic data and the associated ground-truth.

\vspace{-0.3cm}
\paragraph{Reading Network}
In the reading step our main goal is to localize 3 landmarks: the start marker, the end marker, and the needle tip (see the blue dots in \Cref{fig:model_pipeline} (g)). The ground-truth for these targets is provided in the synthetic dataset as angles between 0 to 359. We parameterize all three labels as the angle created between the respective gauge landmark and a horizontal line in the middle of the image, which we define as a 0$\degree$ point (see the dashed pink lines \Cref{fig:model_pipeline} (g)). 

We jointly model all these prediction targets with a single convolutional backbone augmented with three independent prediction heads (one head per prediction target). Although this task could be approached as a regression problem (\textit{i.e} predict a number between 0 to 359), we favor the empirical stability of the gradients created by the cross-entropy loss and approach this problem as a classification task (\textit{i.e} select a class between 0 to 359). Therefore, we quantize the ground-truth values into 360 bins (one bin representing one degree) and optimize the task as a classification problem. To this end, we chose the MobilnetV2\cite{sandler2018mobilenetv2} or EfficientNet \cite{tan2019efficientnet} as the backbones and complement each encoder with three linear heads, each head has an output size of $1 \times 360$. 

\begin{figure*}[t]
    \centering
    \includegraphics[width=\linewidth]{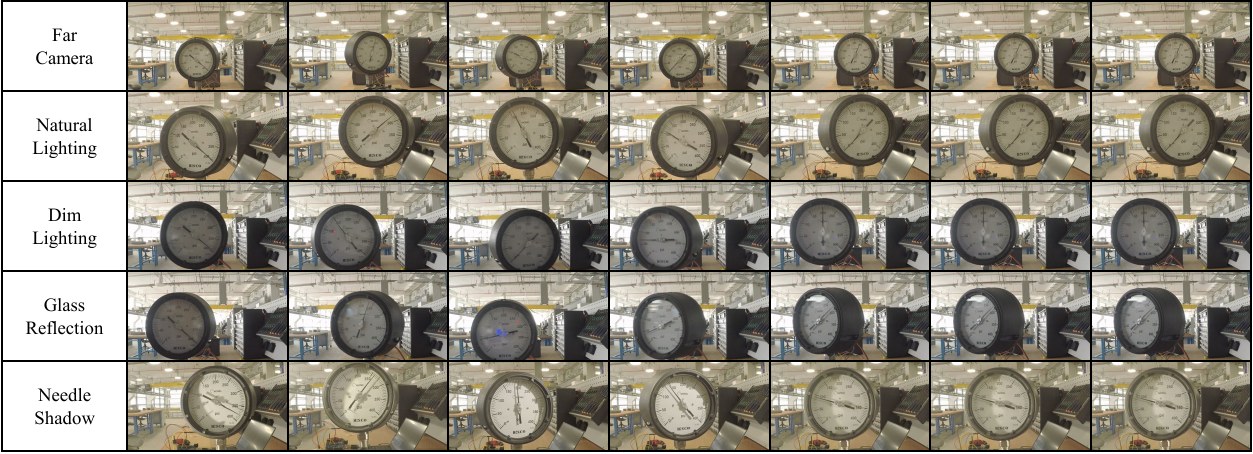}
    \caption{\textbf{Landmark Detection Dataset.} We show some samples of the Landmark detection dataset, each image has an associated ground truth that contains the needle angle, and the location of the start and end of the sticker. We create this dataset using a mechanical arm where we control the relative position of the gauge and the camera, along with the location of the pointer needle. By synchronizing the motion of the arm and the servomotors with the capture device, we can obtain an exact ground truth reading for every frame.}
    \label{fig:lab_samples}
    \vspace{-0.2cm}
\end{figure*}

At training time, we leverage the location prior provided by the segmentation step by appending the segmentation result to the input image in the channel dimension. This segmentation provides a form of visual guidance that allows us to direct the network’s attention toward the gauge instead of the background. We extend the filters on the first layer of each backbone to accept 4 channel inputs, the new filter data is initially set to the average of the original 3 channel filter. Additional training details are provided in the \textbf{supplementary material}.

\section{Gauge Reading Datasets}
\label{sec:datasets}

In this section, we outline our second dataset contribution, namely two real-world datasets that contain manually curated ground-truth labels. The first one is the \textit{land-mark dataset} which contains labels for the location of the start and end markers, along with the pointer needle orientation on 4.813 real-world images. The landmark dataset serves as a held-out test-set to validate our proposed approach. The second dataset provides the semantic segmentation ground-truth for 59 real-world images, it serves as a validation set for the semantic segmentation step.

\subsection{Landmark Regression Dataset}
\label{subsec:lab_dataset}

The landmark dataset consists of 5 video sequences captured using a GoPro camera and an analog gauge attached to the end of a Novint Falcon haptic controller, which was adapted to serve as a robotic arm. This mechanism enables us to control the location of the gauge in the Y-axis (up and down the camera plane) and Z-axis (away or close to the camera plane). This allows us to simulate the capture of gauge images from diverse points of view and varying distances. All the image data is captured and encoded as a video sequence at 720p ($1280 \times 720)$ resolution. 

In addition to the capture viewpoint, the gauge is manipulated with two additional servomotors controlled by an Arduino Uno microcontroller. The first servomotor controls the orientation of the pointer needle allowing us to control the actual gauge reading. As the servomotor is not directly connected to the needle, but rather is attached to the internal mechanical structure of the gauge. The mapping of the servo motor rotation and the needle's effective rotation is non-linear. To approximate this mapping, we sampled and manually recorded the amount of motor rotation input required for the needle to match a specific reading (tick marker) on the gauge's face. Then we linearly interpolate the motor inputs to approximate intermediate angles. The second servomotor can rotate the gauge's face relative to the camera plane (the gauge face can be oriented towards the left or right of the Field of View, which represents the yaw rotation). 

To account for some real-world capture conditions and noise, 5 different environments were simulated. These scenarios include varying light sources and camera locations. In total, each scenario contains between 960 to 966 images. A few samples from the landmark dataset in each of the 5 proposed scenarios are presented in \Cref{fig:lab_samples}.

\vspace{-0.3cm}
\paragraph{Natural Lighting.} Serves as the baseline scenario without any complex conditions. We arrange some artificial lighting that approximates an outdoor setup by daylight and locate the camera close to the gauge (30 cm away). As a result, the case of the gauge occupies on average 30\% of the image's area. Clearly, the gauge is the main object in the image, but there is some background clutter on the capture.

\vspace{-0.3cm}
\paragraph{Far Camera.} Simulates the model's performance when reading slightly more distant gauges. These images were captured 50 cm away from the gauge. On average, the gauge's case occupies 10\% of the image area, which means the global image features are dominated by the background.

\vspace{-0.3cm}
\paragraph{Dim Light.} In this scenario we turn off the artificial lighting, such that the capture condition simulates the use of the gauge reading tool indoors, or with diminished sunlight. The reduced lighting paired with the relatively high (25 fps) video capture introduces some noticeable motion blur artifacts in this scenario, making the location of the landmarks more challenging. 

\vspace{-0.3cm}
\paragraph{Glass Reflection.} This scenario was captured with an open window allowing for direct sunlight towards the front of the gauge, additional objects were added to the scene to create reflections that could alter the appearance of the gauge.

\vspace{-0.3cm}
\paragraph{Needle Shadow.} For this scenario, we set the artificial light source much closer to the gauge and arrange the lightning direction at an angle (never fully frontal). This setup creates sharp shadows in the gauge sticker, mainly from the gauge case and the needle. These conditions often lead to erroneous estimations, as automatic reading models tend to miss-classify the needle's shadow as the actual needle. 

All the gauge readings in the landmark dataset are controlled by the servomotors and carefully synchronized with the video capture. Therefore, our landmark dataset offers a standard benchmark with accurate ground truth labels that enables the validation (and potentially training) of automatic gauge reading tools. Overall, the landmark dataset presents a diverse set of possible failure cases to test against, although it has a bias regarding the gauge type as the same gauge was used for all 5 capture scenarios. 

\begin{figure*}[t]
    \centering
    \includegraphics[width=\textwidth]{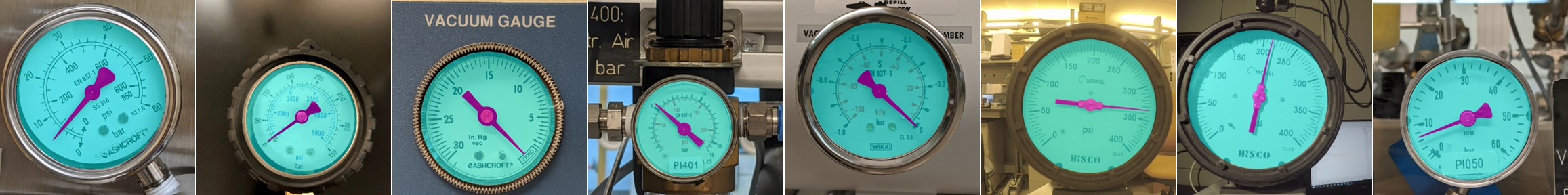}
    \caption{\textbf{Segmented Gauge Dataset.} We provide a dataset with 59 manually segmented real-world gauges. This dataset includes multiple indoor and outdoor gauges and supports the validation of the attention step in our pipeline. In this figure the Gauge pixels are highlighted in Bright cyan, while the pointer needle pixels are shown in magenta, we do not highlight the background pixels for easier visualization.} 
    \label{fig:seg_samples}
    \vspace{-0.2cm}
\end{figure*}

\subsection{Segmented Gauge Dataset}
\label{subsec:seg_dataset}
Our proposed approach relies on image segmentation of the background, gauge, and pointer needle. To validate the performance of our methods on this initial task, we manually segmented 59 real images. Of these 59 images, 43 are part of the Landmark Dataset, the remaining 16 were obtained from other gauges inside an industrial facility. The label distribution in this dataset is highly imbalanced, with the background, needle, and gauge classes representing 66.26\%, 1.03\%, and 32.71\% of the pixels, respectively. We provide this dataset as the segmentation of gauges is a key step in our approach. In addition, it will also support future automatic reading methods that build upon a location step. We visualize some samples of the segmentations contained in this dataset in \Cref{fig:seg_samples}.

\section{Results and Analysis}
\label{sec:results}

We now perform the empirical evaluation of our method, we begin with a direct comparison against the state-of-the-art in the test-set proposed by Howells \textit{et al.} \cite{Howells_2021_CVPR}. Then, we provide the evaluation of our approach in the proposed Landmark dataset. We conclude by assessing the effectiveness of the attention component using our segmentation dataset. 

\subsection{Comparison against the State-of-the-art}
\label{subsec:angle_benchmark}

The work of Howells \etal \cite{Howells_2021_CVPR} provides its own real-world angle reading dataset. We run our proposed approach in their test-set and perform a direct comparison. Their test-set consists of 6 visually distinct gauges, for each individual gauge there are 3 video clips containing a total of 450 frames (150 frames per video clip, 2.700 images total). Each clip shows a static capture (i.e. no camera or gauge motion, no illumination changes) of the gauge where only the pointer needle changes its orientation. The associated angle ground-truth is available for every clip. 

The gauge reading dataset in \cite{Howells_2021_CVPR} was deliberately collected with small gauges, occupying about 10-15\% of the image, such that reading pipelines first crop out the target gauge and then regress the needle's location. Since the semantic segmentation step in our pipeline works as an attention mechanism rather than a detection step, we use OpenCV's implementation of the Hough Transform circle detection \cite{opencv_library, 4767964} to extract an initial crop of the image. We keep the default settings for the algorithm and directly apply our method to the cropped detections. \Cref{tab:benchmark_results} summarizes the results of our approach in the test-set of \cite{Howells_2021_CVPR}.

\begin{table}[t]
    \centering
    \begin{tabular}{l r r r}
        \toprule
        \multirow{2}{*}{\textbf{Dataset}} & \multicolumn{2}{c}{\textbf{Ours}} & \\ \cline{2-3}

        & \textbf{Small } 
        & \textbf{Large} 
        & \textbf{Howells \etal} \\ 
        
        \midrule
              
        \color{lightgray} meter\_a  & \color{lightgray} 19.33   &  \color{lightgray} 16.62   & \color{lightgray} \textbf{9.24} \\
        meter\_a (rev.)      & 5.69          &  \textbf{2.50}    & --- \\ 
        meter\_b            & \textbf{8.23} & 8.29              & 26.06 \\ 
        meter\_c            & 2.74          & 1.97              & \textbf{1.94} \\ 
        meter\_d            & 7.68          & \textbf{7.36}     & 11.70 \\ 
        meter\_e            & 4.32          & 4.83             & \textbf{3.70} \\ 
        meter\_f            & 4.64          & 4.69              & \textbf{4.31} \\
        
        \midrule
        
        \textbf{Average}        & 7.82 & \textbf{7.30}  & 9.49 \\
        \textbf{Average (rev.)}  & 5.55 & \textbf{4.94}  & --- \\ 
        
        \bottomrule
    
    \end{tabular}
    \caption{\textbf{State-of-the-art Comparison.} We test our pipeline in the test-set of \cite{Howells_2021_CVPR}. We report the mean absolute error of the predicted gauge angle readings in degrees. Our small model uses MobileNetV2, which is the same model used by \cite{Howells_2021_CVPR}. The large model uses the EfficientNet-B2 backbone.}
    \label{tab:benchmark_results}
    \vspace{-0.2cm}
\end{table}

Overall, we observe that our proposal outperforms the state-of-the-art by at least 1.67$^{\circ}$, which represents a relative improvement of 17\%. Although \cite{Howells_2021_CVPR} outperforms our method in 3 sequences (c, e, f), we observe that the results are nearly identical in those scenarios as their absolute differences are 0.03$^{\circ}$, 0.62$^{\circ}$, and 0.33$^{\circ}$  respectively. Meanwhile, our method largely outperforms their baseline in the more challenging sequences (b and d), reducing the error by 17.83$^{\circ}$  (68.4\% relative improvement) in sequence b, and 4.34$^{\circ}$  (37.1\% relative error reduction). 

\begin{figure}[t]
    \centering
     \includegraphics[width=\linewidth]{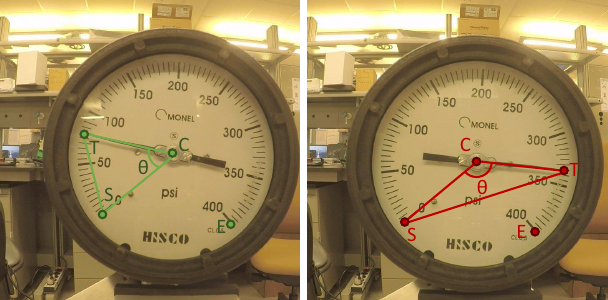}
    \caption{\textbf{Benchmark Ground-Truth Discrepancies.} The angle extraction method used in \cite{Howells_2021_CVPR} extracts the explement angle between the start marker and needle tip when the angle exceeds 180$^{\circ}$. The left figure shows angles for which both methods succeed. The right figure shows an apparent failure case of our proposed model, given the design of the ground-truth labels.}
    \label{fig:angle_error_howells}
    \vspace{-0.2cm}
\end{figure}

\vspace{-0.35cm}
\paragraph{Ground-truth and Predictions Design} We note that the test-set of \cite{Howells_2021_CVPR} follows a different approach to our training set. While we use every angle from 0$^{\circ}$  to 360$^{\circ}$  in our labels and prediction heads, Howells \etal use labels and predictions between 0$^{\circ}$  to 180$^{\circ}$. Their angle prediction formula from four key-points always takes the smallest of the two possible angles between the start marker and needle tip. A visualization of this discrepancy is presented in \Cref{fig:angle_error_howells}. 

Our model's classification head with angles in the range $[0^{\circ}$, 360$^{\circ}$ ) overcomes the aforementioned limitation. Moreover, the same four key-point method proposed by \cite{Howells_2021_CVPR} can be modified to incorporate the sign of the 2D cross-product of the central edges $\overrightarrow{CS} \times \overrightarrow{CT}$ to determine if the angle $\theta$ should be used or its explement ($360^{\circ}  - \theta$). We correct for this discrepancy on every ground-truth angle that is above 180$^{\circ}$. Additional benchmark ground-truth discussion is provided in the \textbf{supplementary material}.

\vspace{-0.3cm}
\paragraph{Ground-truth Correction} In \Cref{tab:benchmark_results} we include two different evaluations for meter\_a, we identify that the sequence meter\_a, contains flawed ground-truth data that does not match the actual reading of the gauge. We manually inspect the entire sequence and identify that this annotation error appears on 40 contiguous frames of the first video capture and on the entire 150 frames of the second. We detail the corrections made in the \textbf{supplementary material}.

\begin{figure}[t]
     \includegraphics[width=0.95\linewidth]{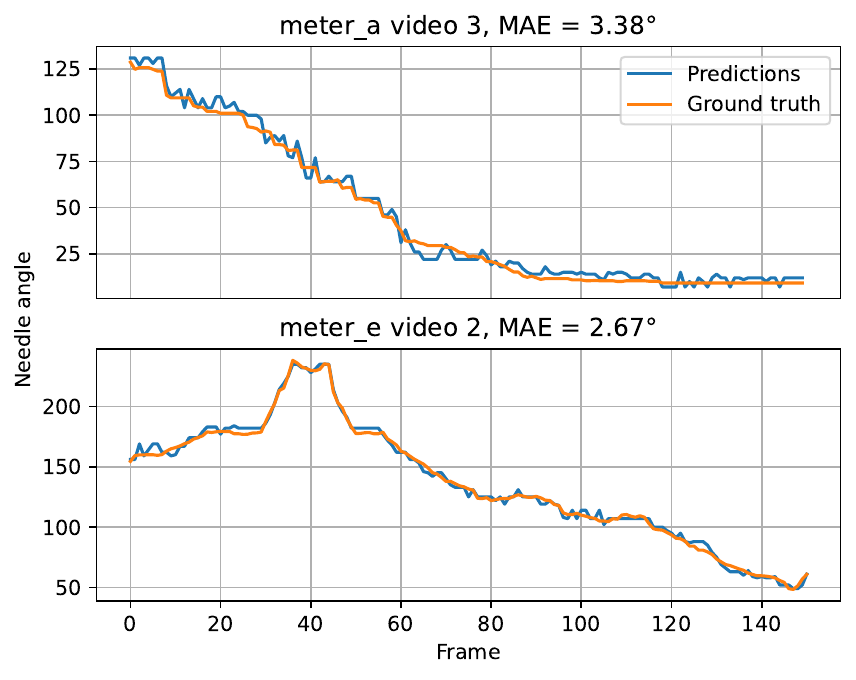}
    \caption{\textbf{Benchmark Time Series Prediction Samples.} Our small (MobileNetV2) model's performance on two sample videos from the benchmark dataset. Our model's needle angle prediction closely matches the ground-truth. All the time series results are presented in the supplementary material.}
    \label{fig:benchmark_timeseries_samples}
    \vspace{-0.2cm}
\end{figure}

We create a revisited (rev) version of meter\_a, with ground-truth values that match the needle location. For completeness, we report the performance of our method on both the full sequence and the revisited subset. We observe a significant gap when the revisited data (rev) is used. Furthermore, we also break down the average performance according to the selected sequence\_a data (original or revisited). With the revisited data, our approach increases the performance gap against the state-of-the-art by up to 4.55$^{\circ}$  in the average error (52\% relative improvement). \Cref{fig:benchmark_timeseries_samples} shows our model's performance on two videos of the meter\_a, video 3 uses the original ground truth, while video2 uses our revisited version. In the \textbf{supplementary material}, we include a similar analysis for every video in the dataset.

\subsection{Landmark Dataset Evaluation}
We now assess the effectiveness of our pipeline in the landmark dataset. Unlike the test-set of \cite{Howells_2021_CVPR}, we directly apply our method to the test data, \textit{i.e.} we do not perform any additional detection step or image cropping. For this evaluation, we also test how our synthetic dataset enables the fine-tuning of models deeper than MobileNetV2 \cite{sandler2018mobilenetv2}. To this end, we choose to fine-tune the EfficienNet \cite{tan2019efficientnet}, \Cref{tab:landmark_performance} summarizes the results.

\begin{table}[b]
    \small
    \centering
    \begin{tabular}{l| r r r r r}
    \toprule
     \multicolumn{1}{c|}{\multirow{2}{*}{\textbf{Scenario}}} &
     \multicolumn{1}{c}{\textbf{Mobile-}}
     & \multicolumn{4}{c}{\textbf{EfficientNet}} \\ \cline{3-6} 
     & \multicolumn{1}{c}{\textbf{NetV2}}
     & \multicolumn{1}{c}{\textbf{B0}}
     & \multicolumn{1}{c}{\textbf{B1}}
     & \multicolumn{1}{c}{\textbf{B2}}
     & \multicolumn{1}{c}{\textbf{B3}} \\ 

    \midrule
    Natural Lighting    & 2.83 & 2.62 & 2.59 & \textbf{2.29} & 2.31 \\
    Far Camera          & 5.15 & 2.81 & 2.93 & \textbf{2.48} & 2.52 \\
    Dim Light           & 3.62 & 5.50 & 4.56 & \textbf{3.20} & 4.10 \\ 
    Glass Reflection    & 2.88 & 2.60 & 2.64 & \textbf{2.37} & 2.44 \\ 
    Needle Shadow       & 3.15 & 3.36 & \textbf{2.85} & 3.11 & 3.14 \\ 

    \midrule    
    \textbf{Average}    & 3.53 & 3.38 & 3.11 & \textbf{2.69} & 2.90 \\ 
    \bottomrule
        
    \end{tabular}
    \caption{\textbf{Performance on the Landmark Dataset.} We evaluate our proposed pipeline on the Landmark Dataset using MobilnetV2 and various sizes of the EfficientNet. The cell values represent the mean absolute angular errors in degrees. The best results are obtained with the EfficientNet B2, nevertheless, the other models remain close in terms of absolute distance.}
    \label{tab:landmark_performance}
    \vspace{-0.2cm}
\end{table}

The best results are obtained with the EfficientNet B2 Network, which corresponds to our Large model in \Cref{tab:benchmark_results}. Although the MobilnetV2 (small model) underperforms the large model in every scenario, the small model shows competitive results in the presence of shadows or low light. Our benchmark shows that the baseline scenario (Natural Lighting) reports the best results, whereas our pipeline reduces its performance in the presence of low light or strong shadows.

We observe that the EfficientNet family can be fine-tuned with our synthetic data up to the B2 size without showing overfit. The larger B3 model does not report any empirical improvement in any scenario. Our best model reports an average angular distance across all scenarios of 2.69$^{\circ}$. 

\begin{figure*}[ht]
    \centering
    \includegraphics[width=\linewidth]{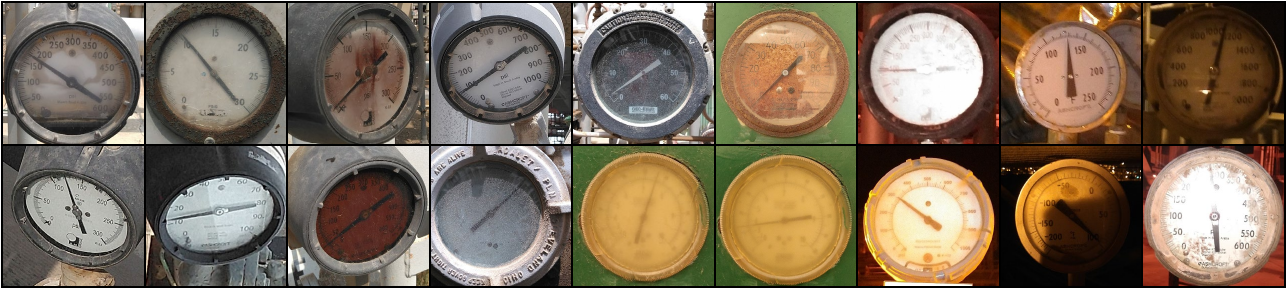}
    \caption{\textbf{Qualitative Field Results.} Field samples were used for field sample analysis in \Cref{subsec:field_dataset} showing our models' performance on difficult gauges with odd view angles, harsh lighting conditions, and various obstructions. Using the EfficientNet-B2 model, the field gauge images were correctly predicted within 2\% of the gauge's range or two marker ticks of the ground-truth.} 
    \label{fig:field_dataset}
    \vspace{-0.2cm}
\end{figure*}

\subsection{Qualitative Analysis}
\label{subsec:field_dataset}

We tested our large (EfficientNet-B2) and small (MobileNetV2) models on real-world gauges located in an industrial facility. These gauges are mostly outdoors and exhibit various signs of environmental degradation. \Cref{fig:field_dataset} shows a set of 18 field gauges that were captured under various visual conditions such as extreme brightness, low lighting, reflections, glare, etc. Many gauge glasses were dusty, tinted, cracked, or opaque due to sunlight exposure. Some of the gauges were not easily accessible and were captured from a distance or from non-frontal view angles.

For this assessment, we include the gauge min and max values, and manually estimate the gauge actual reading (not only the angular locations). Our EfficientNet-B2 model was able to successfully predict all the field samples in \Cref{fig:field_dataset} with an error of less than 2\% of the gauge's range or two marker ticks of the ground-truth, while the MobileNetV2 model predicted 14 out of the 18 sample images correctly under the same tolerance criterion. The range percentage and marker ticks criterion is commonly used in the field instead of the angular error, it is faster and easier to compute and compare. The EfficientNet-B2 and MobileNetV2 models have achieved $2.3^{\circ}$ and $4.46^{\circ}$ MAE respectively on this sample of field images.

\subsection{Segmentation Results}

\begin{table}[t]
    \centering
    \begin{tabular}{l | r r} 
        \toprule
        \multicolumn{1}{c|}{\multirow{2}{*}{\textbf{Class}}}
        & \multicolumn{1}{c}{\textbf{Percentage}}
        & \multicolumn{1}{c}{\textbf{Mean}} \\
        & \multicolumn{1}{c}{\textbf{of Dataset}}
        & \multicolumn{1}{c}{\textbf{IoU}} \\
        \midrule
        
        Background &      66.26 \% & 94.60 \% \\
        Gauge Case &      32.71 \% & 88.13 \% \\ 
        Needle Pointer &  1.03 \% & 76.28 \% \\ 
        \bottomrule
    \end{tabular}
    \caption{\textbf{Segmentation Performance.} Intersection over the union of our method on the segmented dataset from \Cref{subsec:seg_dataset}. The segmentation classes are highly imbalanced, with the needle class accounting only for just 1\% of the labeled pixels. Consequently, this is the class that is harder to segment.}
    \label{tab:segmentation}
    \vspace{-0.2cm}
\end{table}

Finally, we validate the segmentation step using our real-world segmented dataset from \Cref{subsec:seg_dataset}. The evaluation of the segmentation performance is presented in \Cref{tab:segmentation}. Overall, the segmentation network achieves a very good result at locating the gauge case, with an IoU over 94\% IoU for the background class, and the case IoU just below 89\%. Although the needle pointer also shows satisfactory performance, it is the lowest among the 3 classes at 76.3\%.

\begin{figure}[t]
     \includegraphics[width=0.95\linewidth]{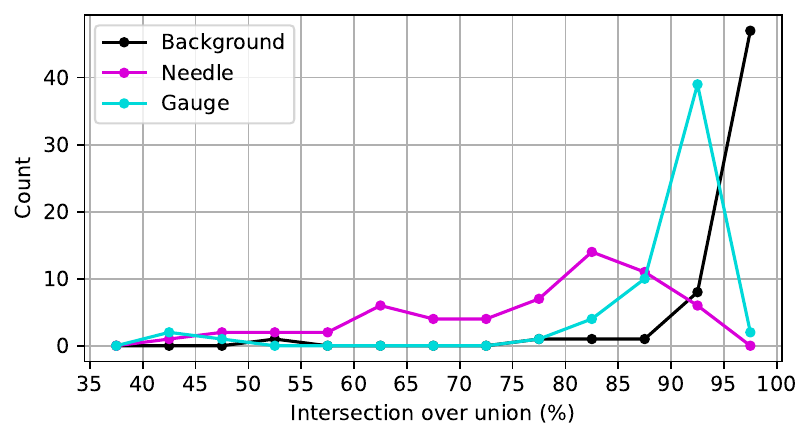}
    \caption{\textbf{Segmentation Model Results.} Distribution of segmentation model IoU results on real images from the segmented dataset in \Cref{subsec:seg_dataset}. IoU values are binned into 5\% ranges, i.e. 90-95\%. Naturally, the sparse nature of the needle class resulted in the poorest results.}
    \label{fig:seg_iou}
    \vspace{-0.2cm}
\end{figure}

We complement this result with the analysis of the error distribution for each individual class in \Cref{fig:seg_iou}. The gauge case (cyan line) and the background show that most of the images report an IoU over 85\% for these two classes. However, the needle has a long tail that extends to just above 55\%. Although the needle performance is lower in comparison to the other 2 classes, most of the images obtain an approximate segmentation of the needle pointer.

\section{Conclusion}
\label{sec:conclusion}

We have introduced a simple yet effective pipeline for the automatic gauge reading task. By means of a two-stage CNN network that is trained on synthetic data. Our proposal is validated on two real-world datasets showing that synthetic data and augmentations can effectively bridge the domain gap. Our method provides a robust technique to read and directly digitize analog gauges, showing strong performance even in challenging and noisy real-world scenarios.


{\small
\bibliographystyle{template/ieee_fullname}
\bibliography{GaugeReading}
}

\clearpage
\section{Implementation Details}
\label{sec:implementation_details}

We implement the segmentation network using the MMsegmentation toolbox \cite{mmseg2020} running over the PyTorch framework \cite{paszke2017automatic}, we trained the segmentation network for 80,000 iterations using a learning rate of $1e^{-7}$ and a batch size of 18. We initialize the segmentation network with the pre-trained weights of the Cityscapes dataset \cite{cordts2016cityscapes} and train until convergence on a held-out set of 500 synthetic images. We empirically find that augmentation plays a key role in closing the domain gap between synthetic (training set) and real-world images (validation sets), therefore we alter the image contrast, brightness, and saturation, we also apply random rotations up to 20\degree, introduce Gaussian blur and randomly crop out small squares in the input image. We train with an input image size of 512$\times$512.

For the reading network, we train the MobilNetV2 backbone with for 120 epochs using a learning rate of $1e^{-7}$ and perform a learning rate step every 50 epochs, we also include a dropout layer just before the classifier and set the dropout probability to 0.4. For the EfficentNet-  B2 model we use a learning rate of $1e^{-7}$ and a dropout of 0.5. For both networks we use the ImageNet pre-trained weights, and include a similar augmentation strategy to the one proposed on the segmentation, but remove the rotations as we observe no empirical improvement. We train the EfficientNet B2 model to convergence in about 6 hours using a single NVIDIA-V100 GPU.

\section{Ablation Analysis}

We present an ablation analysis on two elements in our proposed pipeline, we evaluate the effectiveness of the network if the attention provided by the segmentation is removed and if we drop all the augmentation from the training process. We perform this ablation analysis in the landmark dataset, \cref{tab:ablation} summarizes these results. Overall, we observe both scenarios generate a drop in performance to about 6.1\degree MAE, with the dim light scenario showing the largest performance degradation, but also the far camera setting shows a relative loss in performance of  40\% and 38\%.

\begin{table}[t]
    \small
    \centering
    \begin{tabular}{l| r r r }
    \toprule
    
    Scenario & Large & No Seg. & No Aug. \\
 
    \midrule
    Natural Lighting    &\textbf{2.29} & 3.49   & 3.29 \\
    Far Camera          &\textbf{2.48} & 4.90   & 4.77 \\
    Dim Light           &\textbf{3.20} & 14.48  & 14.45 \\ 
    Glass Reflection    &\textbf{2.37} & 3.71   & 4.26 \\ 
    Needle Shadow       &\textbf{3.11} & 3.96   & 3.99 \\ 

    \midrule    
    \textbf{Average}    & \textbf{2.69} & 6.11 & 6.15 \\ 
    \bottomrule
        
    \end{tabular}
    \caption{\textbf{Ablation Analysis}. We show the relevance of two components in our poplin, the segmentation step (No Seg.) and the data augmentation (No Aug.) overall the largest degradation is shown in low light conditions.}
    \label{tab:ablation}
    \vspace{-0.2cm}
\end{table}

\section{Benchmark Ground-truth Corrections}
\label{sec:benchmark_gt_fixes}

As stated on the main paper, we now outline in detail the corrections made to the ground-truth labels of the benchmark dataset provided by \cite{Howells_2021_CVPR}:

\paragraph{Angles above 180\degree}
The test sequences \texttt{mater\_b\_vid2}, \texttt{meter\_e\_vid1}, and \texttt{meter\_3\_vid2} contain angles that exceed 180 degrees. As mentioned in the main paper the ground-truth did not use the clockwise angular distance from start to needle angle. Instead, it calculated the shortest angular distance in any direction (clockwise or counterclockwise).

We noticed that \texttt{vid\_a\_1}'s first few frames are incorrectly labeled, where the needle is exactly on the start, but the ground-truth label is listed around 20\degree. We opted to modify those values to reflect the actual ground-truth  and report the revisited (rev) performance using those new values. For completeness, we also report the performance with the original ground-truth values. All the frames from \texttt{vid\_a\_2} were off by roughly 42\degree from the actual reading. The metadata of that video's ground-truth asserted that the readings were measured from the marker for 2 bar, even though the minimum marker is at roughly 0.5 bar. This caused an almost constant offset between the actual reading and the provided ground-truth. 

We also note that in Howells \etal the MAE for all of \texttt{meter\_a} was less than 10\degree, however the (nearly constant) 42\degree offset error for all frames of video 2 (a third of total frames) should result in about 14\degree total error for a model with perfect angle readings in the first and third video. However, the results of \cite{Howells_2021_CVPR} are well below this estimation (9.24\degree). We believe that the published ground-truth could be different (or perhaps an older version) from the one used in their experimental setup.

The \texttt{delta\_a} value for \texttt{meter\_e}, which specifies the angle between the start and end of the markers, was corrected from roughly 150\degree to 300\degree. This does not affect Howells \etal or our model, but perhaps future works which rely strongly on this angular distance could be affected.

\section{Detailed Benchmark Results}
\label{sec:full_results}

\begin{figure*}[!htb]
    \centering
    \includegraphics[width=\linewidth]{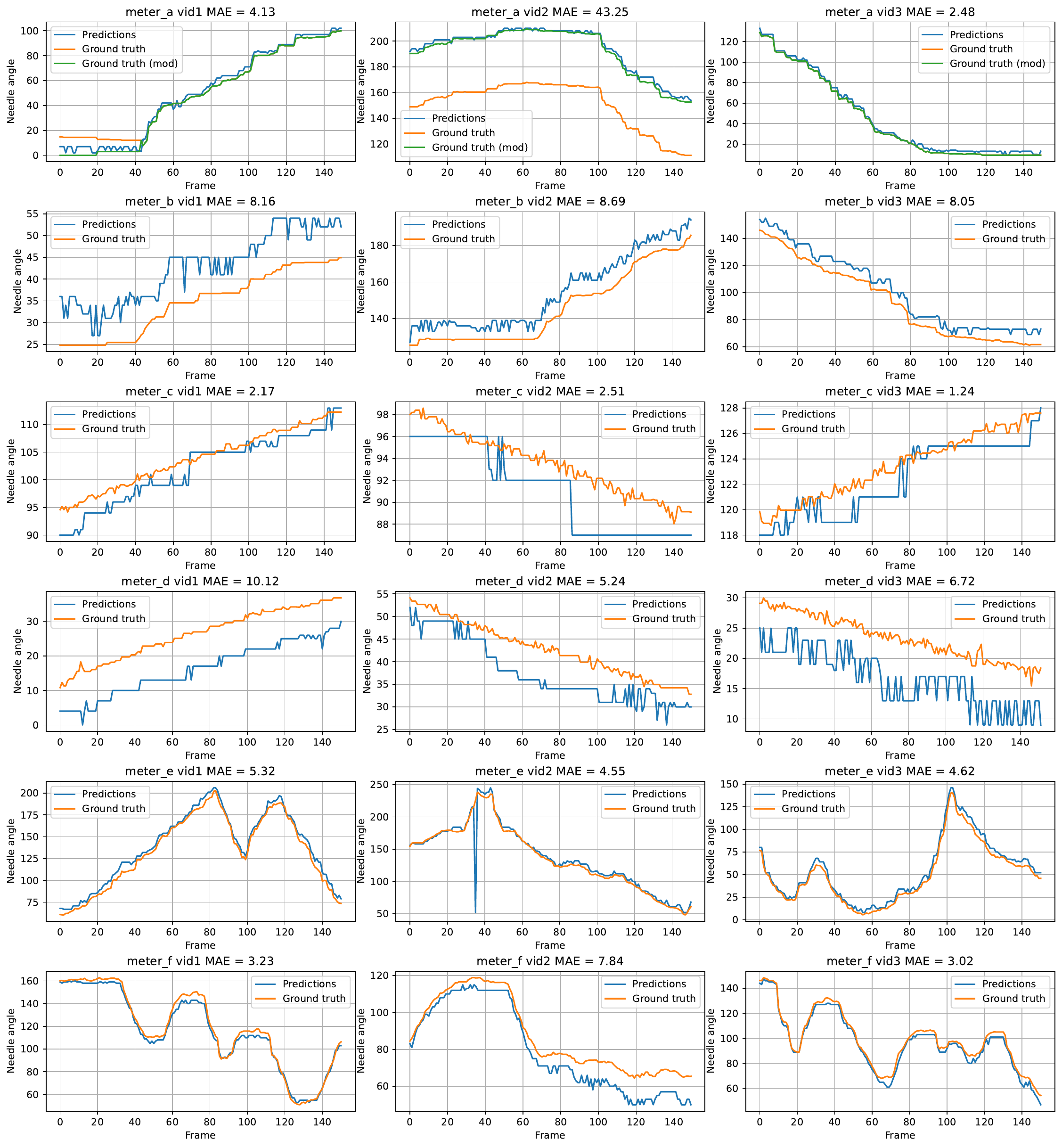}
    \caption{Full benchmark results for the EfficientNet-B2 model. The modified ground-truth values for \texttt{meter\_a} are also shown. The MAE values for \texttt{meter\_a} are with respect to the unmodified ground-truth.}
    \label{fig:benchmark_results_full_eb2}
\end{figure*}

\begin{figure*}[!htb]
    \centering
    \includegraphics[width=\linewidth]{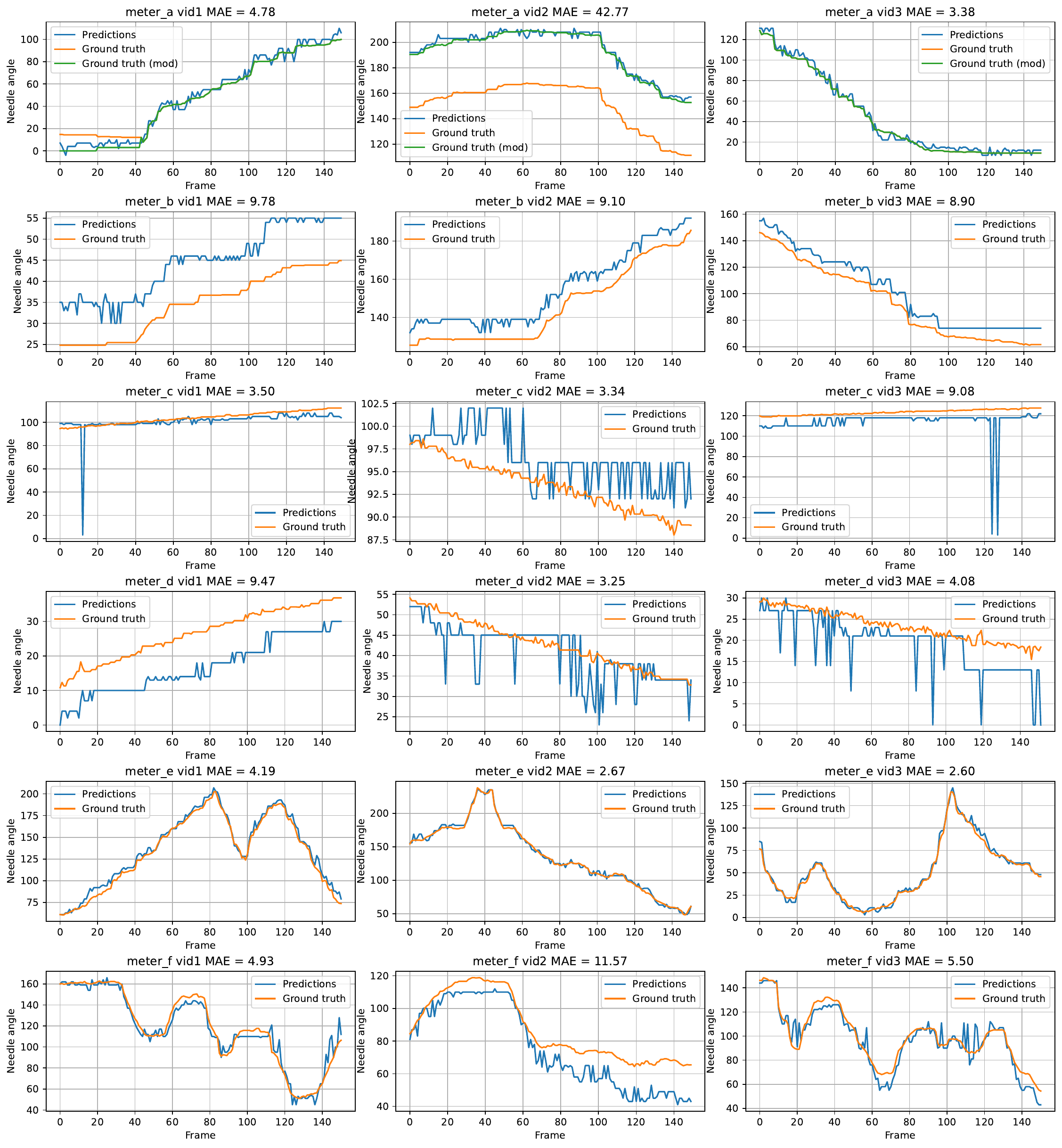}
    \caption{Full benchmark results for the MobileNetV2 model. The modified ground-truth values for \texttt{meter\_a} are also shown. The MAE values for \texttt{meter\_a} are with respect to the unmodified ground-truth.}
    \label{fig:benchmark_results_full_mobilenet}
\end{figure*}

Full benchmark results for the large EfficientNet-B2 model are shown in \cref{fig:benchmark_results_full_eb2} and the results for the small MobileNetV2 model are illustrated in \cref{fig:benchmark_results_full_mobilenet}

\end{document}